# Conditional Random Fields as Recurrent Neural Networks

Shuai Zheng*[1], Sadeep Jayasumana*[1], Bernardino Romera-Paredes[1], Vibhav Vineet†[1,2], Zhizhong Su[3], Dalong Du[3], Chang Huang[3], and Philip H. S. Torr[1]

[1]University of Oxford [2]Stanford University [3]Baidu Institute of Deep Learning

## Abstract

*Pixel-level labelling tasks, such as semantic segmentation, play a central role in image understanding. Recent approaches have attempted to harness the capabilities of deep learning techniques for image recognition to tackle pixel-level labelling tasks. One central issue in this methodology is the limited capacity of deep learning techniques to delineate visual objects. To solve this problem, we introduce a new form of convolutional neural network that combines the strengths of Convolutional Neural Networks (CNNs) and Conditional Random Fields (CRFs)-based probabilistic graphical modelling. To this end, we formulate mean-field approximate inference for the Conditional Random Fields with Gaussian pairwise potentials as Recurrent Neural Networks. This network, called CRF-RNN, is then plugged in as a part of a CNN to obtain a deep network that has desirable properties of both CNNs and CRFs. Importantly, our system fully integrates CRF modelling with CNNs, making it possible to train the whole deep network end-to-end with the usual back-propagation algorithm, avoiding offline post-processing methods for object delineation.*

*We apply the proposed method to the problem of semantic image segmentation, obtaining top results on the challenging Pascal VOC 2012 segmentation benchmark.*

## 1. Introduction

Low-level computer vision problems such as semantic image segmentation or depth estimation often involve assigning a label to each pixel in an image. While the feature representation used to classify individual pixels plays an important role in this task, it is similarly important to consider factors such as image edges, appearance consistency and spatial consistency while assigning labels in order to obtain accurate and precise results.

Designing a strong feature representation is a key challenge in pixel-level labelling problems. Work on this topic includes: TextonBoost [52], TextonForest [51], and Random Forest-based classifiers [50]. Recently, supervised deep learning approaches such as large-scale deep Convolutional Neural Networks (CNNs) have been immensely successful in many high-level computer vision tasks such as image recognition [31] and object detection [20]. This motivates exploring the use of CNNs for pixel-level labelling problems. The key insight is to learn a strong feature representation end-to-end for the pixel-level labelling task instead of hand-crafting features with heuristic parameter tuning. In fact, a number of recent approaches including the particularly interesting works FCN [37] and DeepLab [10] have shown a significant accuracy boost by adapting state-of-the-art CNN based image classifiers to the semantic segmentation problem.

However, there are significant challenges in adapting CNNs designed for high level computer vision tasks such as object recognition to pixel-level labelling tasks. Firstly, traditional CNNs have convolutional filters with large receptive fields and hence produce coarse outputs when restructured to produce pixel-level labels [37]. Presence of max-pooling layers in CNNs further reduces the chance of getting a fine segmentation output [10]. This, for instance, can result in non-sharp boundaries and blob-like shapes in semantic segmentation tasks. Secondly, CNNs lack smoothness constraints that encourage label agreement between similar pixels, and spatial and appearance consistency of the labelling output. Lack of such smoothness constraints can result in poor object delineation and small spurious regions in the segmentation output [59, 58, 32, 39].

On a separate track to the progress of deep learning techniques, probabilistic graphical models have been developed as effective methods to enhance the accuracy of pixel-level labelling tasks. In particular, Markov Random Fields (MRFs) and its variant Conditional Random Fields (CRFs) have observed widespread success in this area [32, 29] and have become one of the most successful graphical models used in computer vision. The key idea of CRF inference for semantic labelling is to formulate the label assignment

*Authors contributed equally.
†Work conducted while authors at the University of Oxford.

problem as a probabilistic inference problem that incorporates assumptions such as the label agreement between similar pixels. CRF inference is able to refine weak and coarse pixel-level label predictions to produce sharp boundaries and fine-grained segmentations. Therefore, intuitively, CRFs can be used to overcome the drawbacks in utilizing CNNs for pixel-level labelling tasks.

One way to utilize CRFs to improve the semantic labelling results produced by a CNN is to apply CRF inference as a post-processing step disconnected from the training of the CNN [10]. Arguably, this does not fully harness the strength of CRFs since it is not integrated with the deep network. In this setup, the deep network is unaware of the CRF during the training phase.

In this paper, we propose an end-to-end deep learning solution for the pixel-level semantic image segmentation problem. Our formulation combines the strengths of both CNNs and CRF based graphical models in one unified framework. More specifically, we formulate mean-field approximate inference for the dense CRF with Gaussian pairwise potentials as a Recurrent Neural Network (RNN) which can refine coarse outputs from a traditional CNN in the forward pass, while passing error differentials back to the CNN during training. Importantly, with our formulation, the whole deep network, which comprises a traditional CNN and an RNN for CRF inference, can be trained end-to-end utilizing the usual back-propagation algorithm.

Arguably, when properly trained, the proposed network should outperform a system where CRF inference is applied as a post-processing method on independent pixel-level predictions produced by a pre-trained CNN. Our experimental evaluation confirms that this indeed is the case. We evaluate the performance of our network on the popular Pascal VOC 2012 benchmark, achieving a new state-of-the-art accuracy of 74.7%. Our source code and models are publicly available [1].

## 2. Related Work

In this section we review approaches that make use of deep learning and CNNs for low-level computer vision tasks, with a focus on semantic image segmentation. A wide variety of approaches have been proposed to tackle the semantic image segmentation task using deep learning. These approaches can be categorized into two main strategies.

The first strategy is based on utilizing separate mechanisms for feature extraction, and image segmentation exploiting the edges of the image [2, 38]. One representative instance of this scheme is the application of a CNN for the extraction of meaningful features, and using superpixels to account for the structural pattern of the image. Two representative examples are [19, 38], where the authors first obtained superpixels from the image and then used a feature extraction process on each of them. The main disadvantage of this strategy is that errors in the initial proposals (e.g: super-pixels) may lead to poor predictions, no matter how good the feature extraction process is. Pinheiro and Collobert [46] employed an RNN to model the spatial dependencies during scene parsing. In contrast to their approach, we show that a typical graphical model such as a CRF can be formulated as an RNN to form a part of a deep network, to perform end-to-end training combined with a CNN.

The second strategy is to directly learn a nonlinear model from the images to the label map. This, for example, was shown in [17], where the authors replaced the last fully connected layers of a CNN by convolutional layers to keep spatial information. An important contribution in this direction is [37], where Long *et al.* used the concept of fully convolutional networks, and the notion that top layers obtain meaningful features for object recognition whereas low layers keep information about the structure of the image, such as edges. In their work, connections from early layers to later layers were used to combine these cues. Bell *et al.* [5] and Chen *et al.* [10, 41] used a CRF to refine segmentation results obtained from a CNN. Bell *et al.* focused on material recognition and segmentation, whereas Chen *et al.* reported very significant improvements on semantic image segmentation. In contrast to these works, which employed CRF inference as a standalone post-processing step disconnected from the CNN training, our approach is an end-to-end trainable network that jointly learns the parameters of the CNN and the CRF in one unified deep network.

Works that use neural networks to predict structured output are found in different domains. For example, Do *et al.* [14] proposed an approach to combine deep neural networks and Markov networks for sequence labeling tasks. Jain *et al.* [26] has shown Convolutional Neural Networks can perform well like MRFs/CRFs approaches in image restoration application. Another domain which benefits from the combination of CNNs and structured loss is handwriting recognition. In natural language processing, Yao *et al.* [60] shows that the performance of an RNN-based words tagger can be significantly improved by incorporating elements of the CRF model. In [6], the authors combined a CNN with Hidden Markov Models for that purpose, whereas more recently, Peng *et al.* [45] used a modified version of CRFs. Related to this line of works, in [25] a joint CNN and CRF model was used for text recognition on natural images. Tompson *et al.* [57] showed the use of joint training of a CNN and an MRF for human pose estimation, while Chen *et al.* [11] focused on the image classification problem with a similar approach. Another prominent work is [21], in which the authors express deformable part models, a kind of MRF, as a layer in a neural network. In our approach, we cast a different graphical model as a neural

[1] https://github.com/torrvision/crfasrnn

network layer.

A number of approaches have been proposed for automatic learning of graphical model parameters and joint training of classifiers and graphical models. Barbu *et al.* [4] proposed a joint training of a MRF/CRF model together with an inference algorithm in their Active Random Field approach. Domke [15] advocated back-propagation based parameter optimization in graphical models when approximate inference methods such as mean-field and belief propagation are used. This idea was utilized in [28], where a binary dense CRF was used for human pose estimation. Similarly, Ross *et al.* [47] and Stoyanov *et al.* [54] showed how back-propagation through belief propagation can be used to optimize model parameters. Ross *et al.* [21], in particular proposes an approach based on learning messages. Many of these ideas can be traced back to [55], which proposes unrolling message passing algorithms as simpler operations that could be performed within a CNN. In a different setup, Krähenbühl and Koltun [30] demonstrated automatic parameter tuning of dense CRF when a modified mean-field algorithm is used for inference. An alternative inference approach for dense CRF, not based on mean-field, is proposed in [61].

In contrast to the works described above, our approach shows that it is possible to formulate dense CRF as an RNN so that one can form an end-to-end trainable system for semantic image segmentation which combines the strengths of deep learning and graphical modelling.

After our initial publication of the technical report of this work on arXiv.org, a number of independent works [49, 35] appeared on arXiv.org presenting similar joint training approaches for semantic image segmentation.

## 3. Conditional Random Fields

In this section we provide a brief overview of Conditional Random Fields (CRF) for pixel-wise labelling and introduce the notation used in the paper. A CRF, used in the context of pixel-wise label prediction, models pixel labels as random variables that form a Markov Random Field (MRF) when conditioned upon a global observation. The global observation is usually taken to be the image.

Let $X_i$ be the random variable associated to pixel $i$, which represents the label assigned to the pixel $i$ and can take any value from a pre-defined set of labels $\mathcal{L} = \{l_1, l_2, \ldots, l_L\}$. Let $\mathbf{X}$ be the vector formed by the random variables $X_1, X_2, \ldots, X_N$, where $N$ is the number of pixels in the image. Given a graph $G = (V, E)$, where $V = \{X_1, X_2, \ldots, X_N\}$, and a global observation (image) $\mathbf{I}$, the pair $(\mathbf{I}, \mathbf{X})$ can be modelled as a CRF characterized by a Gibbs distribution of the form $P(\mathbf{X} = \mathbf{x}|\mathbf{I}) = \frac{1}{Z(\mathbf{I})} \exp(-E(\mathbf{x}|\mathbf{I}))$. Here $E(\mathbf{x})$ is called the energy of the configuration $\mathbf{x} \in \mathcal{L}^N$ and $Z(\mathbf{I})$ is the partition function [33]. From now on, we drop the conditioning on $\mathbf{I}$ in the notation for convenience.

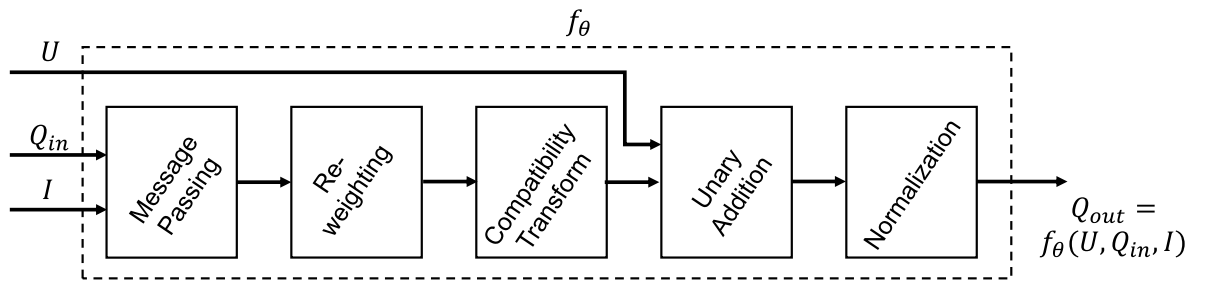

Figure 1. **A mean-field iteration as a CNN.** A single iteration of the mean-field algorithm can be modelled as a stack of common CNN layers.

In the fully connected pairwise CRF model of [29], the energy of a label assignment $\mathbf{x}$ is given by:

$$E(\mathbf{x}) = \sum_i \psi_u(x_i) + \sum_{i<j} \psi_p(x_i, x_j), \qquad (1)$$

where the unary energy components $\psi_u(x_i)$ measure the inverse likelihood (and therefore, the cost) of the pixel $i$ taking the label $x_i$, and pairwise energy components $\psi_p(x_i, x_j)$ measure the cost of assigning labels $x_i, x_j$ to pixels $i, j$ simultaneously. In our model, unary energies are obtained from a CNN, which, roughly speaking, predicts labels for pixels without considering the smoothness and the consistency of the label assignments. The pairwise energies provide an image data-dependent smoothing term that encourages assigning similar labels to pixels with similar properties. As was done in [29], we model pairwise potentials as weighted Gaussians:

$$\psi_p(x_i, x_j) = \mu(x_i, x_j) \sum_{m=1}^{M} w^{(m)} k_G^{(m)}(\mathbf{f}_i, \mathbf{f}_j), \qquad (2)$$

where each $k_G^{(m)}$ for $m = 1, \ldots, M$, is a Gaussian kernel applied on feature vectors. The feature vector of pixel $i$, denoted by $\mathbf{f}_i$, is derived from image features such as spatial location and RGB values [29]. We use the same features as in[29]. The function $\mu(.,.)$, called the label compatibility function, captures the compatibility between different pairs of labels as the name implies.

Minimizing the above CRF energy $E(\mathbf{x})$ yields the most probable label assignment $\mathbf{x}$ for the given image. Since this exact minimization is intractable, a mean-field approximation to the CRF distribution is used for approximate maximum posterior marginal inference. It consists in approximating the CRF distribution $P(\mathbf{X})$ by a simpler distribution $Q(\mathbf{X})$, which can be written as the product of independent marginal distributions, i.e., $Q(\mathbf{X}) = \prod_i Q_i(X_i)$. The steps of the iterative algorithm for approximate mean-field inference and its reformulation as an RNN are discussed next.

**Algorithm 1** Mean-field in dense CRFs [29], broken down to common CNN operations.

---

$Q_i(l) \leftarrow \frac{1}{Z_i} \exp\left(U_i(l)\right)$ for all $i$ ▷ Initialization
**while** not converged **do**
  $\tilde{Q}_i^{(m)}(l) \leftarrow \sum_{j \neq i} k^{(m)}(\mathbf{f}_i, \mathbf{f}_j) Q_j(l)$ for all $m$ ▷ Message Passing
  $\check{Q}_i(l) \leftarrow \sum_m w^{(m)} \tilde{Q}_i^{(m)}(l)$ ▷ Weighting Filter Outputs
  $\hat{Q}_i(l) \leftarrow \sum_{l' \in \mathcal{L}} \mu(l, l') \check{Q}_i(l')$ ▷ Compatibility Transform
  $\breve{Q}_i(l) \leftarrow U_i(l) - \hat{Q}_i(l)$ ▷ Adding Unary Potentials
  $Q_i \leftarrow \frac{1}{Z_i} \exp\left(\breve{Q}_i(l)\right)$ ▷ Normalizing
**end while**

---

## 4. A Mean-field Iteration as a Stack of CNN Layers

A key contribution of this paper is to show that the mean-field CRF inference can be reformulated as a Recurrent Neural Network (RNN). To this end, we first consider individual steps of the mean-field algorithm summarized in Algorithm 1 [29], and describe them as CNN layers. Our contribution is based on the observation that filter-based approximate mean-field inference approach for dense CRFs relies on applying Gaussian spatial and bilateral filters on the mean-field approximates in each iteration. Unlike the standard convolutional layer in a CNN, in which filters are fixed after the training stage, we use edge-preserving Gaussian filters [56, 42], coefficients of which depend on the original spatial and appearance information of the image. These filters have the additional advantages of requiring a smaller set of parameters, despite the filter size being potentially as big as the image.

While reformulating the steps of the inference algorithm as CNN layers, it is essential to be able to calculate error differentials in each layer w.r.t. its inputs in order to be able to back-propagate the error differentials to previous layers during training. We also discuss how to calculate error differentials with respect to the parameters in each layer, enabling their optimization through the back-propagation algorithm. Therefore, in our formulation, CRF parameters such as the weights of the Gaussian kernels and the label compatibility function can also be optimized automatically during the training of the full network.

Once the individual steps of the algorithm are broken down as CNN layers, the full algorithm can then be formulated as an RNN. We explain this in Section 5 after discussing the steps of Algorithm 1 in detail below. In Algorithm 1 and the remainder of this paper, we use $U_i(l)$ to denote the negative of the unary energy introduced in the previous section, i.e., $U_i(l) = -\psi_u(X_i = l)$. In the conventional CRF setting, this input $U_i(l)$ to the mean-field algorithm is obtained from an independent classifier.

### 4.1. Initialization

In the initialization step of the algorithm, the operation $Q_i(l) \leftarrow \frac{1}{Z_i} \exp\left(U_i(l)\right)$, where $Z_i = \sum_l \exp(U_i(l))$, is performed. Note that this is equivalent to applying a softmax function over the unary potentials $U$ across all the labels at each pixel. The softmax function has been extensively used in CNN architectures before and is therefore well known in the deep learning community. This operation does not include any parameters and the error differentials received at the output of the step during back-propagation could be passed down to the unary potential inputs after performing usual backward pass calculations of the softmax transformation.

### 4.2. Message Passing

In the dense CRF formulation, message passing is implemented by applying $M$ Gaussian filters on $Q$ values. Gaussian filter coefficients are derived based on image features such as the pixel locations and RGB values, which reflect how strongly a pixel is related to other pixels. Since the CRF is potentially fully connected, each filter's receptive field spans the whole image, making it infeasible to use a brute-force implementation of the filters. Fortunately, several approximation techniques exist to make computation of high dimensional Gaussian filtering significantly faster. Following [29], we use the Permutohedral lattice implementation [1], which can compute the filter response in $O(N)$ time, where $N$ is the number of pixels of the image [1].

During back-propagation, error derivatives w.r.t. the filter inputs are calculated by sending the error derivatives w.r.t. the filter outputs through the same $M$ Gaussian filters in reverse direction. In terms of permutohedral lattice operations, this can be accomplished by only reversing the order of the separable filters in the blur stage, while building the permutohedral lattice, splatting, and slicing in the same way as in the forward pass. Therefore, back-propagation through this filtering stage can also be performed in $O(N)$ time. Following [29], we use two Gaussian kernels, a spatial kernel and a bilateral kernel. In this work, for simplicity, we keep the bandwidth values of the filters fixed. It is also possible to use multiple spatial and bilateral kernels with different bandwidth values and learn their optimal linear combination.

### 4.3. Weighting Filter Outputs

The next step of the mean-field iteration is taking a weighted sum of the $M$ filter outputs from the previous step, for each class label $l$. When each class label is considered individually, this can be viewed as usual convolution with a $1 \times 1$ filter with $M$ input channels, and one output channel. Since both inputs and the outputs to this step are known during back-propagation, the error derivative w.r.t. the filter weights can be computed, making it possible to automatically learn the filter weights (relative contributions from each Gaussian filter output from the previous stage). Error derivative w.r.t. the inputs can also be computed in the usual manner to pass the error derivatives down to the previous stage. To obtain a higher number of tunable parameters, in contrast to [29], we use independent kernel weights for each class label. The intuition is that the relative importance of the spatial kernel vs the bilateral kernel depends on the visual class. For example, bilateral kernels may have on the one hand a high importance in bicycle detection, because similarity of colours is determinant; on the other hand they may have low importance for TV detection, given that whatever is inside the TV screen may have many different colours.

### 4.4. Compatibility Transform

In the compatibility transform step, outputs from the previous step (denoted by $\check{Q}$ in Algorithm 1) are shared between the labels to a varied extent, depending on the compatibility between these labels. Compatibility between the two labels $l$ and $l'$ is parameterized by the label compatibility function $\mu(l, l')$. The Potts model, given by $\mu(l, l') = [l \neq l']$, where $[.]$ is the Iverson bracket, assigns a fixed penalty if different labels are assigned to pixels with similar properties. A limitation of this model is that it assigns the same penalty for all different pairs of labels. Intuitively, better results can be obtained by taking the compatibility between different label pairs into account and penalizing the assignments accordingly. For example, assigning labels "person" and "bicycle" to nearby pixels should have a lesser penalty than assigning labels "sky" and "bicycle". Therefore, learning the function $\mu$ from data is preferred to fixing it in advance with Potts model. We also relax our compatibility transform model by assuming that $\mu(l, l') \neq \mu(l', l)$ in general.

Compatibility transform step can be viewed as another convolution layer where the spatial receptive field of the filter is $1 \times 1$, and the number of input and output channels are both $L$. Learning the weights of this filter is equivalent to learning the label compatibility function $\mu$. Transferring error differentials from the output of this step to the input can be done since this step is a usual convolution operation.

### 4.5. Adding Unary Potentials

In this step, the output from the compatibility transform stage is subtracted element-wise from the unary inputs $U$. While no parameters are involved in this step, transferring error differentials can be done trivially by copying the differentials at the output of this step to both inputs with the appropriate sign.

### 4.6. Normalization

Finally, the normalization step of the iteration can be considered as another softmax operation with no parameters. Differentials at the output of this step can be passed on to the input using the softmax operation's backward pass.

## 5. The End-to-end Trainable Network

We now describe our end-to-end deep learning system for semantic image segmentation. To pave the way for this, we first explain how repeated mean-field iterations can be organized as an RNN.

### 5.1. CRF as RNN

In the previous section, it was shown that one iteration of the mean-field algorithm can be formulated as a stack of common CNN layers (see Fig. 1). We use the function $f_{\boldsymbol{\theta}}$ to denote the transformation done by one mean-field iteration: given an image $I$, pixel-wise unary potential values $U$ and an estimation of marginal probabilities $Q_{\text{in}}$ from the previous iteration, the next estimation of marginal distributions after one mean-field iteration is given by $f_{\boldsymbol{\theta}}(U, Q_{\text{in}}, I)$. The vector $\boldsymbol{\theta} = \{w^{(m)}, \mu(l, l^{'})\}, m \in \{1, ..., M\}, l, l^{'} \in \{l_1, ..., l_L\}$ represents the CRF parameters described in Section 4.

Multiple mean-field iterations can be implemented by repeating the above stack of layers in such a way that each iteration takes $Q$ value estimates from the previous iteration and the unary values in their original form. This is equivalent to treating the iterative mean-field inference as a Recurrent Neural Network (RNN) as shown in Fig. 2. Using the notation in the figure, the behaviour of the network is given by the following equations where $T$ is the number of mean-field iterations:

$$H_1(t) = \begin{cases} \text{softmax}(U), & t = 0 \\ H_2(t-1), & 0 < t \leq T, \end{cases} \tag{3}$$

$$H_2(t) = f_{\boldsymbol{\theta}}(U, H_1(t), I), \quad 0 \leq t \leq T, \tag{4}$$

$$Y(t) = \begin{cases} 0, & 0 \leq t < T \\ H_2(t), & t = T. \end{cases} \tag{5}$$

We name this RNN structure CRF-RNN. Parameters of the CRF-RNN are the same as the mean-field parameters described in Section 4 and denoted by $\boldsymbol{\theta}$ here. Since the calculation of error differentials w.r.t. these parameters in a single iteration was described in Section 4, they can be learnt in the RNN setting using the standard back-propagation through time algorithm [48, 40]. It was shown in [29] that the mean-field iterative algorithm for dense CRF converges in less than 10 iterations. Furthermore, in practice, after about 5 iterations, increasing the number of iterations usually does not significantly improve results [29]. Therefore, it does not suffer from the vanishing and exploding gradient problem inherent to deep RNNs [7, 43]. This allows us to use a plain RNN architecture instead of more sophisticated architectures such as LSTMs in our network.

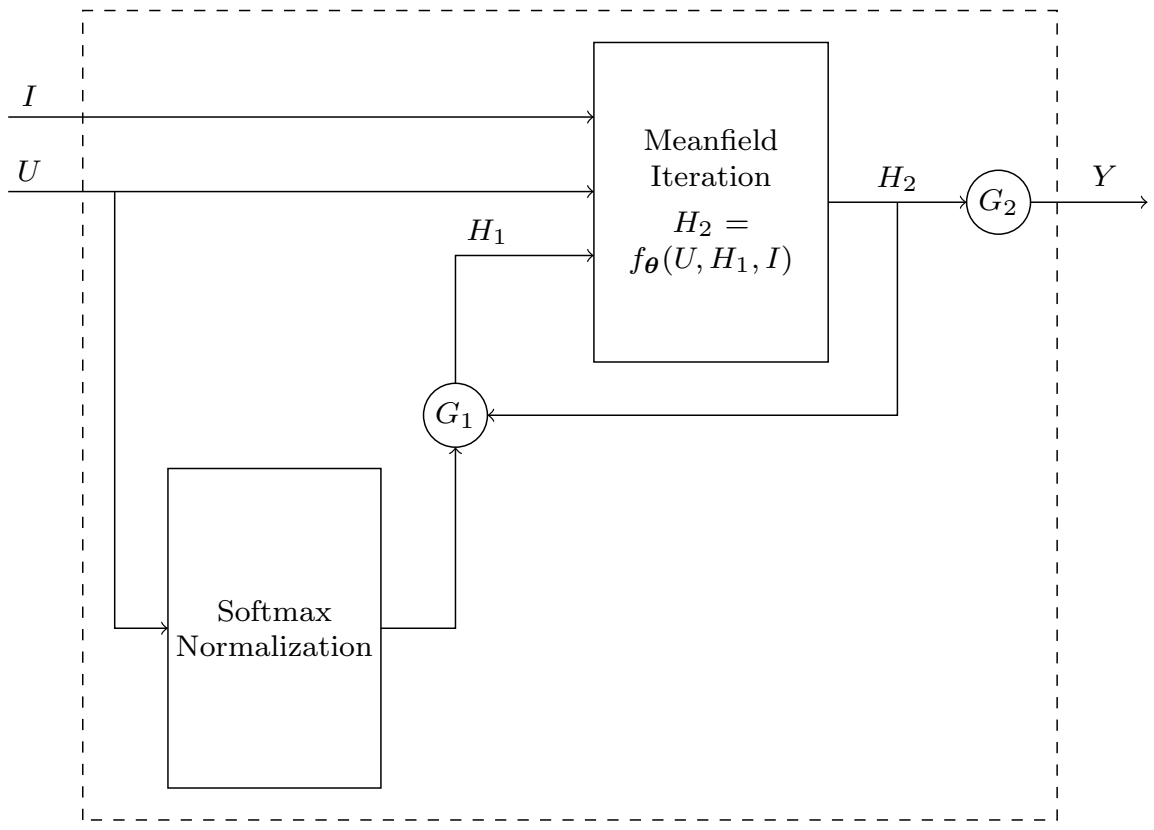

Figure 2. **The CRF-RNN Network.** We formulate the iterative mean-field algorithm as a Recurrent Neural Network (RNN). Gating functions $G_1$ and $G_2$ are fixed as described in the text.

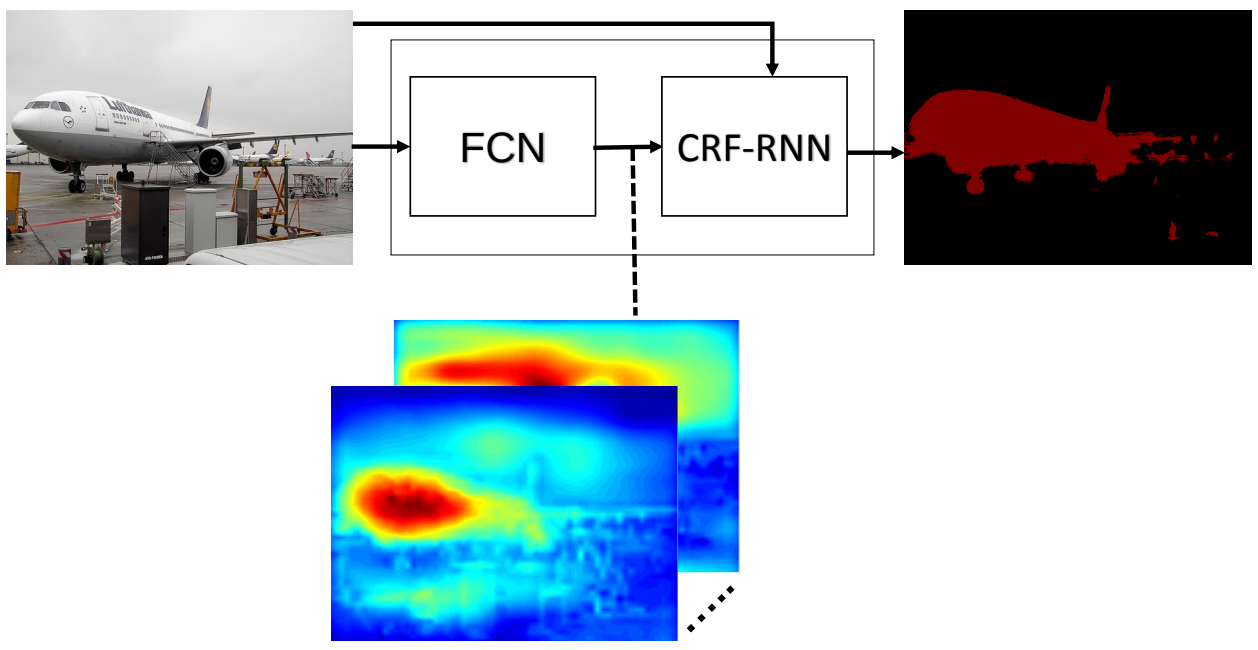

Figure 3. **The End-to-end Trainable Network.** Schematic visualization of our full network which consists of a CNN and the CNN-CRF network. Best viewed in colour.

### 5.2. Completing the Picture

Our approach comprises a fully convolutional network stage, which predicts pixel-level labels without considering structure, followed by a CRF-RNN stage, which performs CRF-based probabilistic graphical modelling for structured prediction. The complete system, therefore, unifies strengths of both CNNs and CRFs and is trainable end-to-end using the back-propagation algorithm [34] and the Stochastic Gradient Descent (SGD) procedure. During training, a whole image (or many of them) can be used as the mini-batch and the error at each pixel output of the network can be computed using an appropriate loss function such as the softmax loss with respect to the ground truth segmentation of the image. We used the FCN-8s architecture of [37] as the first part of our network, which provides unary potentials to the CRF. This network is based on the VGG-16 network [53] but has been restructured to perform pixel-wise prediction instead of image classification. The complete architecture of our network, including the FCN-8s part can be found in the appendix.

In the forward pass through the network, once the computation enters the CRF-RNN after passing through the CNN stage, it takes $T$ iterations for the data to leave the loop created by the RNN. Neither the CNN that provides unary values nor the layers after the CRF-RNN (i.e., the loss layers) need to perform any computations during this time since the refinement happens only inside the RNN's loop. Once the output $Y$ leaves the loop, next stages of the deep network after the CRF-RNN can continue the forward pass. In our setup, a softmax loss layer directly follows the CRF-RNN and terminates the network.

During the backward pass, once the error differentials reach the CRF-RNN's output $Y$, they similarly spend $T$ iterations within the loop before reaching the RNN input $U$ in order to propagate to the CNN which provides the unary input. In each iteration inside the loop, error differentials are computed inside each component of the mean-field iteration as described in Section 4. We note that unnecessarily increasing the number of mean-field iterations $T$ could potentially result in the vanishing and exploding gradient problems in the CRF-RNN. We, however, did not experience this problem during our experiments.

## 6. Implementation Details

In the present section we describe the implementation details of the proposed network, as well as its training process. The high-level architecture of our system, which was implemented using the popular Caffe [27] deep learning library, is shown in Fig. 3. Complete architecture of the deep network can be found in the appendix. The full source code and the trained models of our approach will be made publicly available.

We initialized the first part of the network using the publicly available weights of the FCN-8s network [37]. The compatibility transform parameters of the CRF-RNN were initialized using the Potts model, and kernel width and weight parameters were obtained from a cross-validation process. We found that such initialization results in faster convergence of training. During the training phase, parameters of the whole network were optimized end-to-end using the back-propagation algorithm. In particular we used full image training described in [37], with learning rate fixed at $10^{-13}$ and momentum set to $0.99$. These extreme values of the parameters were used since we employed only one image per batch to avoid reaching memory limits of the GPU.

In all our experiments, during training, we set the number of mean-field iterations $T$ in the CRF-RNN to $5$ to avoid vanishing/exploding gradient problems and to reduce the training time. During the test time, iteration count was increased to 10. The effect of this parameter value on the accuracy is discussed in section 7.1.

**Loss function** During the training of the models that achieved the best results reported in this paper, we used the standard softmax loss function, that is, the log-likelihood error function described in [30]. The standard metric used in the Pascal VOC challenge is the average intersection over union (IU), which we also use here to report the results. In our experiments we found that high values of IU on the validation set were associated to low values of the averaged softmax loss, to a large extent. We also tried the robust log-likelihood in [30] as a loss function for CRF-RNN training. However, this did not result in increased accuracy nor faster convergence.

**Normalization techniques** As described in Section 4, we use the exponential function followed by pixel-wise normalization across channels in several stages of the CRF-RNN. Since this operation has a tendency to result in small gradients with respect to the input when the input value is large, we conducted several experiments where we replaced this by a rectifier linear unit (ReLU) operation followed by a normalization across the channels. Our hypothesis was that this approach may approximate the original operation adequately while speeding up the training due to improved gradients. Furthermore, ReLU would induce sparsity on the probability of labels assigned to pixels, implicitly pruning low likelihood configurations, which could have a positive effect. However, this approach did not lead to better results, obtaining 1% IU lower than the original setting performance.

## 7. Experiments

We present experimental results with the proposed CRF-RNN framework. We use these datasets: the Pascal VOC 2012 dataset, and the Pascal Context dataset. We use the Pascal VOC 2012 dataset as it has become the golden standard to comprehensively evaluate any new semantic segmentation approach in comparison to existing methods. We also use the Pascal Context dataset to assess how well our approach performs on a dataset with different characteristics.

### Pascal VOC Datasets

In order to evaluate our approach with existing methods under the same circumstances, we conducted two main experiments with the Pascal VOC 2012 dataset, followed by a qualitative experiment.

In the first experiment, following [37, 38, 41], we used a training set consisted of VOC 2012 training data (1464 images), and training and validation data of [23], which amounts to a total of 11,685 images. After removing the overlapping images between VOC 2012 validation data and this training dataset, we were left with 346 images from the original VOC 2012 validation set to validate our models on. We call this set the reduced validation set in the sequel. Annotations of the VOC 2012 test set, which consists of 1456 images, are not publicly available and hence the final results on the test set were obtained by submitting the results to the Pascal VOC challenge evaluation server [18]. Regardless of the smaller number of images, we found that the relative improvements of the accuracy on our validation set were in good agreement with the test set.

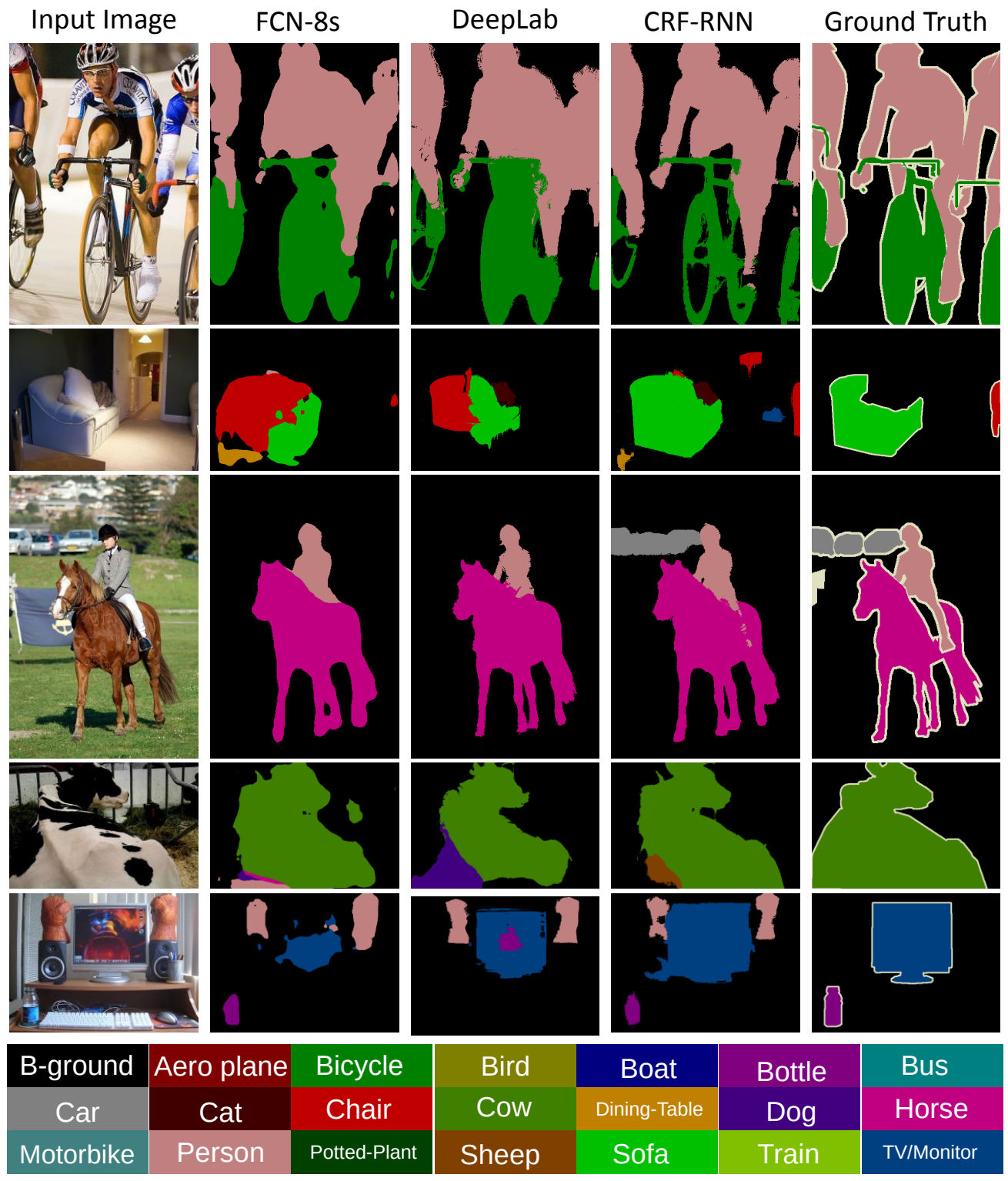

Figure 4. **Qualitative results on the validation set of Pascal VOC 2012.** FCN [37] is a CNN-based model that does not employ CRF. Deeplab [10] is a two-stage approach, where the CNN is trained first, and then CRF is applied on top of the CNN output. Our approach is an end-to-end trained system that integrates both CNN and CRF-RNN in one deep network. Best viewed in colour.

As a first step we directly compared the potential advantage of learning the model end-to-end with respect to alternative learning strategies. These are plain FCN-8s without applying CRF, and with CRF as a postprocessing method disconnected from the training of FCN, which is comparable to the approach described in [10] and [41]. The results are reported in Table 1 and show a clear advantage of the end-to-end strategy over the offline application of CRF as a post-processing method. This can be attributed to the fact that during the SGD training of the CRF-RNN, the CNN component and the CRF component learn how to co-operate with each other to produce the optimum output of the whole network.

We then proceeded to compare our approach with all state-of-the-art methods that used training data from the standard VOC 2012 training and validation sets, and from the dataset published with [22]. The results are shown in Table 2, above the bar, and we can see that our approach outperforms all competitors.

In the second experiment, in addition to the above training set, we used data from the Microsoft COCO dataset [36] as was done in [41] and [12]. We selected images from MS COCO 2014 training set where the ground truth segmentation has at least 200 pixels marked with classes labels present in the VOC 2012 dataset. With this selection, we ended up using 66,099 images from the COCO dataset and therefore a total of 66,099 + 11,685 = 77,784 training images were used in the second experiment. The same reduced validation set was used in this second experiment as well. In this case, we first fine-tuned the plain FCN-32s network (without the CRF-RNN part) on COCO data, then we built an FCN-8s network with the learnt weights and finally train the CRF-RNN network end-to-end using VOC 2012 training data only. Since the MS COCO ground truth segmentation data contains somewhat coarse segmentation masks where objects are not delineated properly, we found that fine-tuning our model with COCO did not yield significant improvements. This can be understood because the primary advantage of our model comes from delineating the objects and improving fine segmentation boundaries. The VOC 2012 training dataset therefore helps our model learn this task effectively. The results of this experiment are shown in Table 2, below the bar, and we see that our approach sets a new state-of-the-art on the VOC 2012 dataset.

Note that in both setups, our approach outperforms competing methods due to the end-to-end training of the CNN and CRF in the unified CRF-RNN framework. We also evaluated our models on the VOC 2010, and VOC 2011 test set (see Table 2). In all cases our method achieves the state-of-the-art performance.

In order to have a qualitative evidence about how CRF-RNN learns, we visualize the compatibility function learned after the training stage of the CRF-RNN as a matrix representation in Fig. 5. Element $(i, j)$ of this matrix corresponds to $\mu(i, j)$ defined earlier: a high value at $(i, j)$ implies high penalty for assigning label $i$ to a pixel when a similar pixel (spatially or appearance wise) is assigned label $j$. For example we can appreciate that the learned compatibility matrix assigns a low penalty to pairs of labels that tend to appear together, such as [*Motorbike*, *Person*], and [*Dining table*, *Chair*].

| Method | Without COCO | With COCO |
|---|---|---|
| Plain FCN-8s | 61.3 | 68.3 |
| FCN-8s and CRF disconnected | 63.7 | 69.5 |
| End-to-end training of CRF-RNN | 69.6 | 72.9 |

Table 1. Mean IU accuracy of our approach, CRF-RNN, compared with similar methods, evaluated on the reduced VOC 2012 validation set.

| Method | VOC 2010 test | VOC 2011 test | VOC 2012 test |
|---|---|---|---|
| BerkeleyRC [3] | n/a | 39.1 | n/a |
| O2PCPMC [8] | 49.6 | 48.8 | 47.8 |
| Divmbest [44] | n/a | n/a | 48.1 |
| NUS-UDS [16] | n/a | n/a | 50.0 |
| SDS [23] | n/a | n/a | 51.6 |
| MSRA-CFM [13] | n/a | n/a | 61.8 |
| FCN-8s [37] | n/a | 62.7 | 62.2 |
| Hypercolumn [24] | n/a | n/a | 62.6 |
| Zoomout [38] | 64.4 | 64.1 | 64.4 |
| Context-Deep-CNN-CRF [35] | n/a | n/a | 70.7 |
| DeepLab-MSc [10] | n/a | n/a | 71.6 |
| **Our method w/o COCO** | 73.6 | 72.4 | 72.0 |
| BoxSup [12] | n/a | n/a | 71.0 |
| DeepLab [10, 41] | n/a | n/a | 72.7 |
| **Our method with COCO** | **75.7** | **75.0** | **74.7** |

Table 2. Mean IU accuracy of our approach, CRF-RNN, compared to the other approaches on the Pascal VOC 2010-2012 test datasets. Methods from the first group do not use MS COCO data for training. The methods from the second group use both COCO and VOC datasets for training.

### Pascal Context Dataset

We conducted an experiment on the Pascal Context dataset [39], which differs from the previous one in the larger number of classes considered, 59. We used the provided partitions of training and validation sets, and the obtained results are reported in Table 3.

| Method | $O_2$P [8] | CFM [13] | FCN-8s [37] | CRF-RNN |
|---|---|---|---|---|
| Mean IU | 18.1 | 34.4 | 37.78 | **39.28** |

Table 3. Mean IU accuracy of our approach, CRF-RNN, evaluated on the Pascal Context validation set.

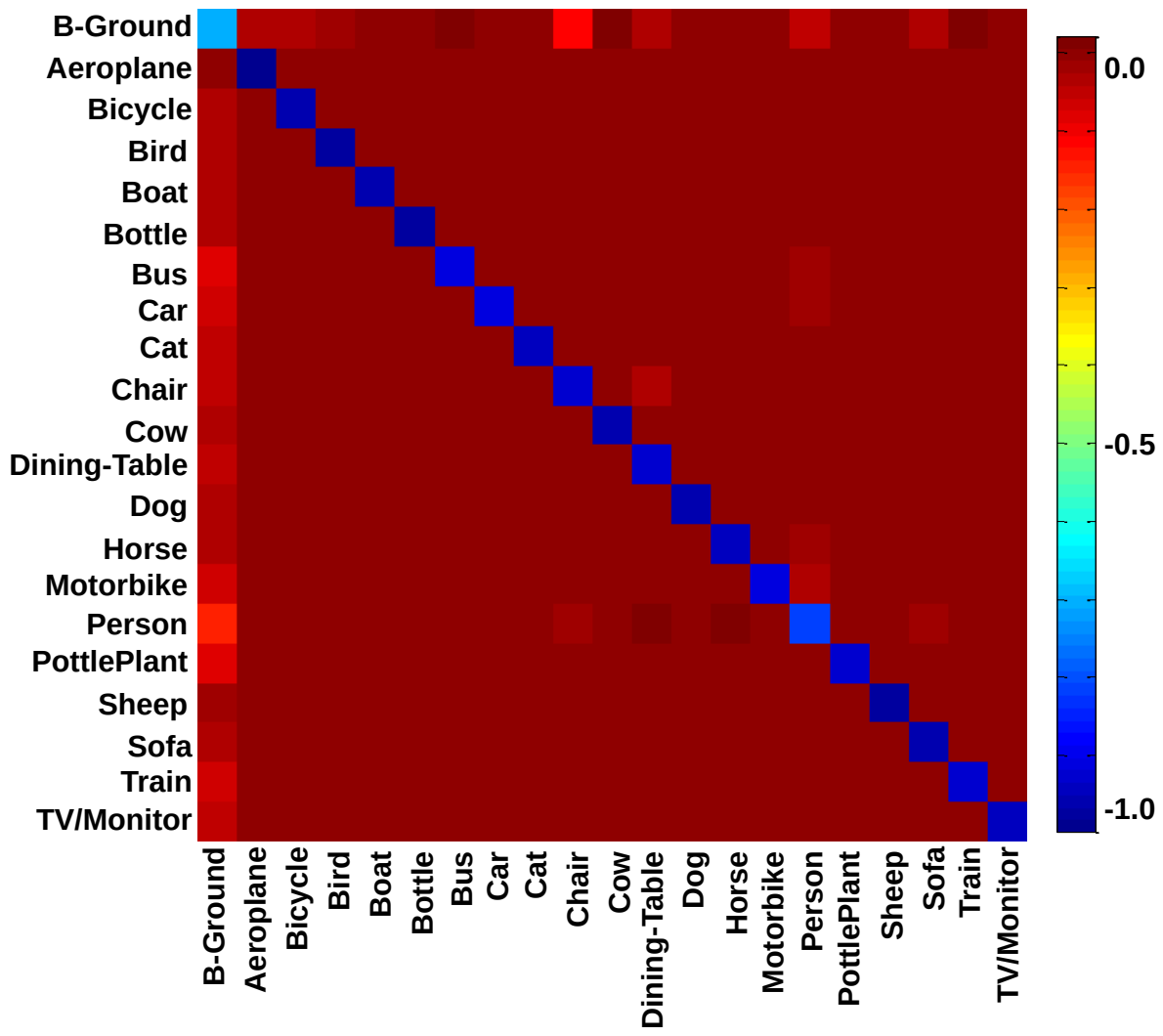

Figure 5. **Visualization of the learnt label compatibility matrix.** In the standard Potts model, diagonal entries are equal to $-1$, while off-diagonal entries are zero. These values have changed after the end-to-end training of our network. Best viewed in colour.

### 7.1. Effect of Design Choices

We performed a number of additional experiments on the Pascal VOC 2012 validation set described above to study the effect of some design choices we made.

We first studied the performance gains attained by our modifications to the CRF over the CRF approach proposed by [29]. We found that using different filter weights for different classes improved the performance by 1.8 percentage points, and that introducing the asymmetric compatibility transform further boosted the performance by 0.9 percentage points.

Regarding the RNN parameter iteration count $T$, incrementing it to $T = 10$ during the test time, from $T = 5$ during the train time, produced an accuracy improvement of 0.2 percentage points. Setting $T = 10$ also during training reduced the accuracy by 0.7 percentage points. We believe that this might be due to a vanishing gradient effect caused by using too many iterations. In practice that leads to the first part of the network (the one producing unary potentials) receiving a very weak error gradient signal during training, thus hampering its learning capacity.

End-to-end training after the initialization of CRF parameters improved performance by 3.4 percentage points. We also conducted an experiment where we froze the FCN-8s part and fine-tuned only the RNN part (i.e., CRF parameters). It improved the performance over initialization by only 1 percentage point. We therefore conclude that end-to-end training significantly contributed to boost the accuracy of the system.

Treating each iteration of mean-field inference as an independent step with its own parameters, and training end-to-end with 5 such iterations yielded a final mean IU score of only 70.9, supporting the hypothesis that the recurrent structure of our approach is important for its success.

## 8. Conclusion

We presented CRF-RNN, an interpretation of dense CRFs as Recurrent Neural Networks. Our formulation fully integrates CRF-based probabilistic graphical modelling with emerging deep learning techniques. In particular, the proposed CRF-RNN can be plugged in as a part of a traditional deep neural network: It is capable of passing on error differentials from its outputs to inputs during back-propagation based training of the deep network while learning CRF parameters. We demonstrate the use of this approach by utilizing it for the semantic segmentation task: we form an end-to-end trainable deep network by combining a fully convolutional neural network with the CRF-RNN. Our system achieves a new state-of-the-art on the popular Pascal VOC segmentation benchmark. This improvement can be attributed to the uniting of the strengths of CNNs and CRFs in a single deep network.

In the future, we plan to investigate the advantages/disadvantages of restricting the capabilities of the RNN part of our network to mean-field inference of dense CRF. A sensible baseline to the work presented here would be to use more standard RNNs (*e.g.* LSTMs) that learn to iteratively improve the input unary potentials to make them closer to the ground-truth.

**Acknowledgement** This work was supported by grants Leverhulme Trust, EPSRC EP/I001107/2 and ERC 321162-HELIOS. We thank the Caffe team, Baidu IDL, and the Oxford ARC team for their support. We gratefully acknowledge GPU donations from NVIDIA.

## References

[1] A. Adams, J. Baek, and M. A. Davis. Fast high-dimensional filtering using the permutohedral lattice. *Computer Graphics Forum*, 29(2):753–762, 2010.

[2] P. Arbeláez, B. Hariharan, C. Gu, S. Gupta, L. Bourdev, and J. Malik. Semantic segmentation using regions and parts. In *IEEE CVPR*, 2012.

[3] P. Arbeláez, M. Maire, C. Fowlkes, and J. Malik. Contour detection and hierarchical image segmentation. *IEEE TPAMI*, 33(5):898–916, 2011.

[4] A. Barbu. Training an active random field for real-time image denoising. *IEEE TIP*, 18(11):2451–2462, 2009.

[5] S. Bell, P. Upchurch, N. Snavely, and K. Bala. Material recognition in the wild with the materials in context database. In *IEEE CVPR*, 2015.

[6] Y. Bengio, Y. LeCun, and D. Henderson. Globally trained handwritten word recognizer using spatial representation, convolutional neural networks, and hidden markov models. In *NIPS*, pages 937–937, 1994.

[7] Y. Bengio, P. Simard, and P. Frasconi. Learning long-term dependencies with gradient descent is difficult. *IEEE Transactions on Neural Networks*, 1994.

[8] J. Carreira, R. Caseiro, J. Batista, and C. Sminchisescu. Free-form region description with second-order pooling. *IEEE TPAMI*, 2014.

[9] K. Chatfield, K. Simonyan, A. Vedaldi, and A. Zisserman. Return of the devil in the details: Delving deep into convolutional nets. In *BMVC*, 2014.

[10] L.-C. Chen, G. Papandreou, I. Kokkinos, K. Murphy, and A. L. Yuille. Semantic image segmentation with deep convolutional nets and fully connected crfs. In *ICLR*, 2015.

[11] L.-C. Chen, A. G. Schwing, A. L. Yuille, and R. Urtasun. Learning deep structured models. In *ICLRW*, 2015.

[12] J. Dai, K. He, and J. Sun. Boxsup: Exploiting bounding boxes to supervise convolutional networks for semantic segmentation. In *arXiv:1503.01640*, 2015.

[13] J. Dai, K. He, and J. Sun. Convolutional feature masking for joint object and stuff segmentation. In *IEEE CVPR*, 2015.

[14] T.-M.-T. Do and T. Artieres. Neural conditional random fields. In *NIPS*, 2010.

[15] J. Domke. Learning graphical model parameters with approximate marginal inference. *IEEE TPAMI*, 35(10):2454–2467, 2013.

[16] J. Dong, Q. Chen, S. Yan, and A. Yuille. Towards unified object detection and semantic segmentation. In *ECCV*, 2014.

[17] D. Eigen, C. Puhrsch, and R. Fergus. Depth map prediction from a single image using a multi-scale deep network. In *NIPS*, 2014.

[18] M. Everingham, S. M. A. Eslami, L. Van Gool, C. K. I. Williams, J. Winn, and A. Zisserman. The pascal visual object classes challenge: A retrospective. *IJCV*, 111(1):98–136, 2015.

[19] C. Farabet, C. Couprie, L. Najman, and Y. LeCun. Learning hierarchical features for scene labeling. *IEEE TPAMI*, 2013.

[20] R. Girshick, J. Donahue, T. Darrell, and J. Malik. Rich feature hierarchies for accurate object detection and semantic segmentation. In *IEEE CVPR*, 2014.

[21] R. Girshick, F. Iandola, T. Darrell, and J. Malik. Deformable part models are convolutional neural networks. In *CVPR*, 2015.

[22] B. Hariharan, P. Arbelaez, L. D. Bourdev, S. Maji, and J. Malik. Semantic contours from inverse detectors. In *IEEE ICCV*, 2011.

[23] B. Hariharan, P. Arbeláez, R. Girshick, and J. Malik. Simultaneous detection and segmentation. In *ECCV*, 2014.

[24] B. Hariharan, P. Arbelaez, R. Girshick, and J. Malik. Hypercolumns for object segmentation and fine-grained localization. In *IEEE CVPR*, 2015.

[25] M. Jaderberg, K. Simonyan, A. Vedaldi, and A. Zisserman. Deep structured output learning for unconstrained text recognition. In *ICLR*, 2015.

[26] V. Jain, J. F. Murray, F. Roth, S. C. Turaga, V. P. Zhigulin, K. L. Briggman, M. Helmstaedter, W. Denk, and H. S. Seung. Supervised learning of image restoration with convolutional networks. In *IEEE ICCV*, 2007.

[27] Y. Jia, E. Shelhamer, J. Donahue, S. Karayev, J. Long, R. Girshick, S. Guadarrama, and T. Darrell. Caffe: Convolutional architecture for fast feature embedding. In *ACM Multimedia*, pages 675–678, 2014.

[28] M. Kiefel and P. V. Gehler. Human pose estmation with fields of parts. In *ECCV*, 2014.

[29] P. Krähenbühl and V. Koltun. Efficient inference in fully connected crfs with gaussian edge potentials. In *NIPS*, 2011.

[30] P. Krähenbühl and V. Koltun. Parameter learning and convergent inference for dense random fields. In *ICML*, 2013.

[31] A. Krizhevsky, I. Sutskever, and G. E. Hinton. Imagenet classification with deep convolutional neural networks. In *NIPS*, 2012.

[32] L. Ladicky, C. Russell, P. Kohli, and P. H. Torr. Associative hierarchical crfs for object class image segmentation. In *IEEE ICCV*, 2009.

[33] J. D. Lafferty, A. McCallum, and F. C. N. Pereira. Conditional random fields: Probabilistic models for segmenting and labeling sequence data. In *ICML*, 2001.

[34] Y. LeCun, L. Bottou, Y. Bengio, and P. Haffner. Gradient-based learning applied to document recognition. *Proceedings of the IEEE*, 86(11):2278–2324, 1998.

[35] G. Lin, C. Shen, I. Reid, and A. van dan Hengel. Efficient piecewise training of deep structured models for semantic segmentation. In *arXiv:1504.01013*, 2015.

[36] T.-Y. Lin, M. Maire, S. Belongie, L. Bourdev, R. Girshick, J. Hays, P. Perona, D. Ramanan, C. L. Zitnick, and P. Dollar. Microsoft coco: Common objects in context. In *arXiv:1405.0312*, 2014.

[37] J. Long, E. Shelhamer, and T. Darrell. Fully convolutional networks for semantic segmentation. In *IEEE CVPR*, 2015.

[38] M. Mostajabi, P. Yadollahpour, and G. Shakhnarovich. Feedforward semantic segmentation with zoom-out features. In *IEEE CVPR*, 2015.

[39] R. Mottaghi, X. Chen, X. Liu, N.-G. Cho, S.-W. Lee, S. Fidler, R. Urtasun, and A. Yuille. The role of context for object detection and semantic segmentation in the wild. In *IEEE CVPR*, 2014.

[40] M. C. Mozer. Backpropagation. chapter A Focused Backpropagation Algorithm for Temporal Pattern Recognition. L. Erlbaum Associates Inc., 1995.

[41] G. Papandreou, L.-C. Chen, K. Murphy, and A. L. Yuille. Weakly- and semi-supervised learning of a dcnn for semantic image segmentation. In *arXiv:1502.02734*, 2015.

[42] S. Paris and F. Durand. A fast approximation of the bilateral filter using a signal processing approach. *IJCV*, 81(1):24–52, 2013.

[43] R. Pascanu, C. Gulcehre, K. Cho, and Y. Bengio. On the difficulty of training recurrent neural networks. In *ICML*, 2013.

[44] G. S. Payman Yadollahpour, Dhruv Batra. Discriminative re-ranking of diverse segmentations. In *IEEE CVPR*, 2013.

[45] J. Peng, L. Bo, and J. Xu. Conditional neural fields. In *NIPS*, 2009.

[46] P. H. O. Pinheiro and R. Collobert. Recurrent convolutional neural networks for scene labeling. In *ICML*, 2014.

[47] S. Ross, D. Munoz, M. Hebert, and J. A. Bagnell. Learning message-passing inference machines for structured prediction. In *IEEE CVPR*, 2011.

[48] D. E. Rumelhart, G. E. Hinton, and R. J. Williams. Neurocomputing: Foundations of research. chapter Learning Internal Representations by Error Propagation. MIT Press, 1988.

[49] A. G. Schwing and R. Urtasun. Fully connected deep structured networks. In *arXiv:1503.02351*, 2015.

[50] J. Shotton, A. Fitzgibbon, M. Cook, T. Sharp, M. Finocchio, R. Moore, A. Kipman, and A. Blake. Real-time human pose recognition in parts from single depth images. In *IEEE CVPR*, 2011.

[51] J. Shotton, M. Johnson, and R. Cipolla. Semantic texton forests for image categorization and segmentation. In *IEEE CVPR*, 2008.

[52] J. Shotton, J. Winn, C. Rother, and A. Criminisi. Textonboost for image understanding: Multi-class object recognition and segmentation by jointly modeling texture, layout, and context. *IJCV*, 81(1):2–23, 2009.

[53] K. Simonyan and A. Zisserman. Very deep convolutional networks for large-scale image recognition. In *arXiv:1409.1556*, 2014.

[54] V. Stoyanov, A. Ropson, and J. Eisner. Empirical risk minimization of graphical model parameters given approximate inference, decoding, and model structure. In *AISTATS*, 2011.

[55] S. C. Tatikonda and M. I. Jordan. Loopy belief propagation and gibbs measures. In *Proceedings of the Eighteenth Conference on Uncertainty in Artificial Intelligence*, 2002.

[56] C. Tomasi and R. Manduchi. Bilateral filtering for gray and color images. In *IEEE CVPR*, 1998.

[57] J. J. Tompson, A. Jain, Y. LeCun, and C. Bregler. Joint training of a convolutional network and a graphical model for human pose estimation. In *NIPS*, 2014.

[58] Z. Tu. Auto-context and its application to high-level vision tasks. In *IEEE CVPR*, 2008.

[59] Z. Tu, X. Chen, A. L. Yuille, and S.-C. Zhu. Image parsing: Unifying segmentation, detection, and recognition. *IJCV*, 63(2):113–140, 2005.

[60] K. Yao, B. Peng, G. Zweig, D. Yu, X. Li, and F. Gao. Recurrent conditional random field for language understanding. In *ICASSP*, 2014.

[61] Y. Zhang and T. Chen. Efficient inference for fully-connected crfs with stationarity. In *CVPR*, 2012.

| Methods trained with COCO | Mean IU | 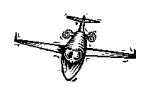 | 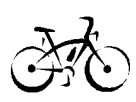 | 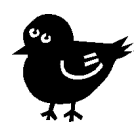 | 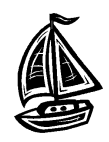 | 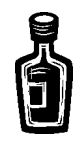 | 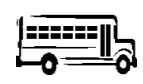 | 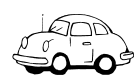 | 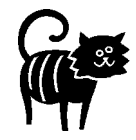 | 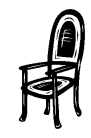 |
|---|---|---|---|---|---|---|---|---|---|---|
| **Our method** | **74.7** | **90.4** | **55.3** | **88.7** | **68.4** | 69.8 | 88.3 | 82.4 | 85.1 | 32.6 |
| DeepLab[10, 41] | 72.7 | 89.1 | 38.3 | 88.1 | 63.3 | 69.7 | 87.1 | 83.1 | 85.0 | 29.3 |
| BoxSup[12] | 71.0 | 86.4 | 35.5 | 79.7 | 65.2 | 65.2 | 84.3 | 78.5 | 83.7 | 30.5 |
| Methods trained w/o COCO | | | | | | | | | | |
| **Our method trained w/o COCO** | 72.0 | 87.5 | 39.0 | 79.7 | 64.2 | 68.3 | 87.6 | 80.8 | 84.4 | 30.4 |
| DeepLab-MSc-CRF-LargeFOV[10] | 71.6 | 84.4 | 54.5 | 81.5 | 63.6 | 65.9 | 85.1 | 79.1 | 83.4 | 30.7 |
| Context_Deep_CNN_CRF[35] | 70.7 | 87.5 | 37.7 | 75.8 | 57.4 | 72.3 | 88.4 | 82.6 | 80.0 | 33.4 |
| Zoomout[38] | 64.4 | 81.9 | 35.1 | 78.2 | 57.4 | 56.5 | 80.5 | 74.0 | 79.8 | 22.4 |
| Hypercolumn[24] | 62.6 | 68.7 | 33.5 | 69.8 | 51.3 | 70.2 | 81.1 | 71.9 | 74.9 | 23.9 |
| FCN-8s[37] | 62.2 | 76.8 | 34.2 | 68.9 | 49.4 | 60.3 | 75.3 | 74.7 | 77.6 | 21.4 |
| MSRA_CFM[13] | 61.8 | 75.7 | 26.7 | 69.5 | 48.8 | 65.6 | 81.0 | 69.2 | 73.3 | 30.0 |
| SDS[23] | 51.6 | 63.3 | 25.7 | 63.0 | 39.8 | 59.2 | 70.9 | 61.4 | 54.9 | 16.8 |
| NUS_UDS [16] | 50.0 | 67.0 | 24.5 | 47.2 | 45.0 | 47.9 | 65.3 | 60.6 | 58.5 | 15.5 |
| TTIC-divmbest-rerank[44] | 48.1 | 62.7 | 25.6 | 46.9 | 43.0 | 54.8 | 58.4 | 58.6 | 55.6 | 14.6 |
| BONN_O2PCPMC_FGT_SEGM [8] | 47.8 | 64.0 | 27.3 | 54.1 | 39.2 | 48.7 | 56.6 | 57.7 | 52.5 | 14.2 |

| Methods trained with COCO | 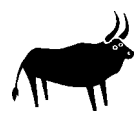 | 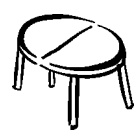 | 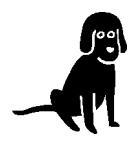 | 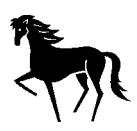 | 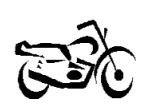 | 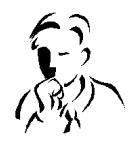 | 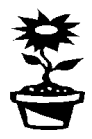 | 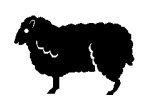 | 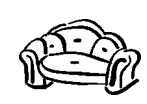 | 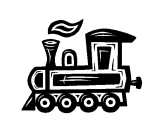 | 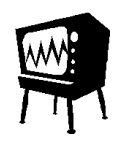 |
|---|---|---|---|---|---|---|---|---|---|---|---|
| **Our method** | **78.5** | **64.4** | 79.6 | **81.9** | **86.4** | 81.8 | 58.6 | 82.4 | 53.5 | 77.4 | **70.1** |
| DeepLab[10, 41] | 76.5 | 56.5 | 79.8 | 77.9 | 85.8 | 82.4 | 57.4 | 84.3 | 54.9 | 80.5 | 64.1 |
| BoxSup[12] | 76.2 | 62.6 | 79.3 | 76.1 | 82.1 | 81.3 | 57.0 | 78.2 | 55.0 | 72.5 | 68.1 |
| Methods trained w/o COCO | | | | | | | | | | | |
| **Our method trained w/o COCO** | 78.2 | 60.4 | 80.5 | 77.8 | 83.1 | 80.6 | 59.5 | 82.8 | 47.8 | 78.3 | 67.1 |
| DeepLab-MSc-CRF-LargeFOV [10] | 74.1 | 59.8 | 79.0 | 76.1 | 83.2 | 80.8 | 59.7 | 82.2 | 50.4 | 73.1 | 63.7 |
| Context_Deep_CNN_CRF[35] | 71.5 | 55.0 | 79.3 | 78.4 | 81.3 | 82.7 | 56.1 | 79.8 | 48.6 | 77.1 | 66.3 |
| TTI_zoomout_16[38] | 69.6 | 53.7 | 74.0 | 76.0 | 76.6 | 68.8 | 44.3 | 70.2 | 40.2 | 68.9 | 55.3 |
| Hypercolumn[24] | 60.6 | 46.9 | 72.1 | 68.3 | 74.5 | 72.9 | 52.6 | 64.4 | 45.4 | 64.9 | 57.4 |
| FCN-8s[37] | 62.5 | 46.8 | 71.8 | 63.9 | 76.5 | 73.9 | 45.2 | 72.4 | 37.4 | 70.9 | 55.1 |
| MSRA_CFM[13] | 68.7 | 51.5 | 69.1 | 68.1 | 71.7 | 67.5 | 50.4 | 66.5 | 44.4 | 58.9 | 53.5 |
| SDS[23] | 45.0 | 48.2 | 50.5 | 51.0 | 57.7 | 63.3 | 31.8 | 58.7 | 31.2 | 55.7 | 48.5 |
| NUS_UDS[16] | 50.8 | 37.4 | 45.8 | 59.9 | 62.0 | 52.7 | 40.8 | 48.2 | 36.8 | 53.1 | 45.6 |
| TTIC-divmbest-rerank[44] | 47.5 | 31.2 | 44.7 | 51.0 | 60.9 | 53.5 | 36.6 | 50.9 | 30.1 | 50.2 | 46.8 |
| BONN_O2PCPMC_FGT_SEGM[8] | 54.8 | 29.6 | 42.2 | 58.0 | 54.8 | 50.2 | 36.6 | 58.6 | 31.6 | 48.4 | 38.6 |

Table 4. Intersection over Union (IU) accuracy of our approach, CRF-RNN, compared to the other state-of-the-art approaches on the Pascal VOC 2012 test set. Scores for other methods were taken the results published by the original authors. The symbols are from Chatfield *et al*. [9].

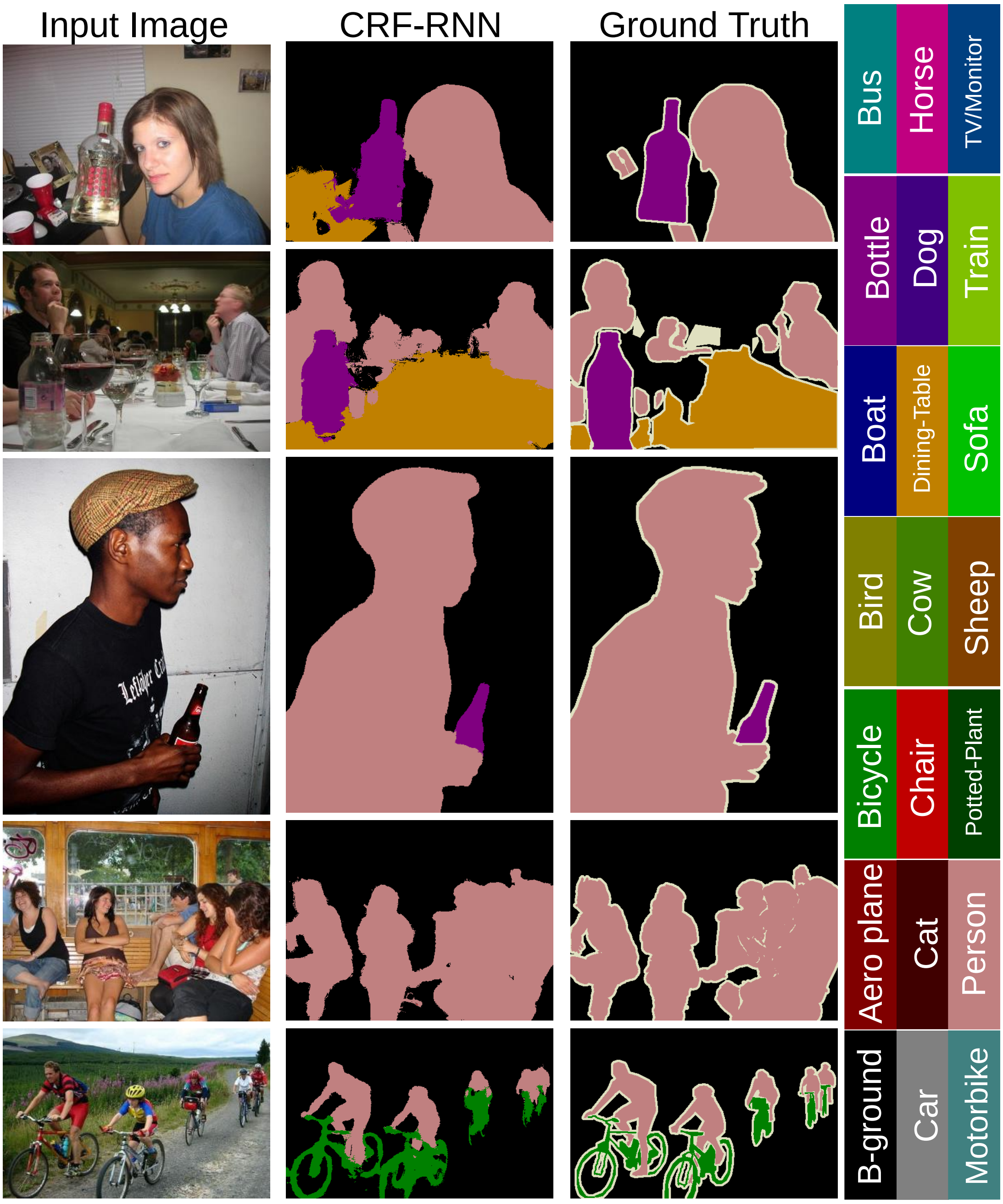

Figure 6. **Typical good quality segmentation results I.** Illustration of sample results on the validation set of the Pascal VOC 2012 dataset. Note that in some cases our method is able to pick correct segmentations that are not marked correctly in the ground truth. Best viewed in colour.

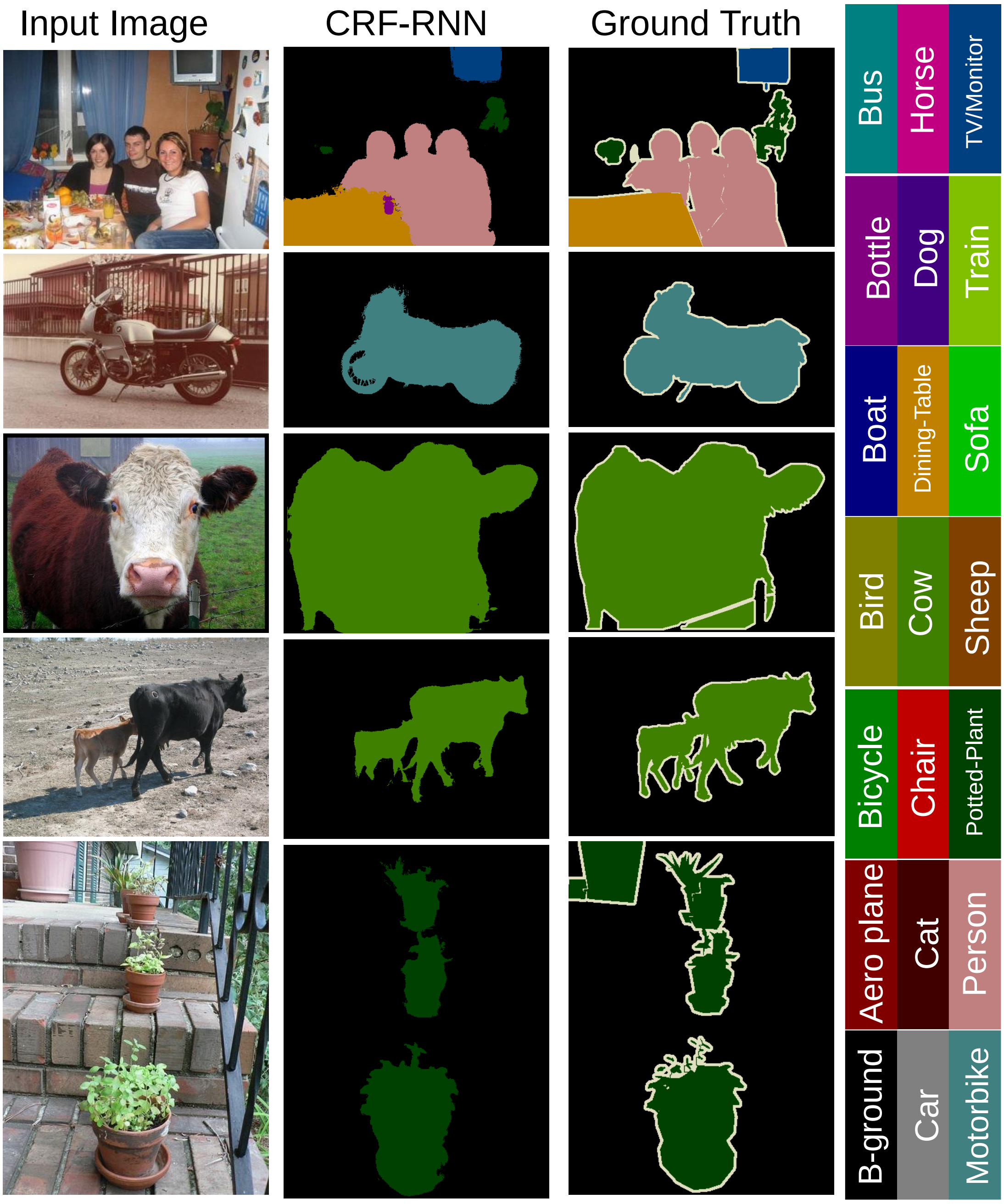

Figure 7. **Typical good quality segmentation results II.** Illustration of sample results on the validation set of the Pascal VOC 2012 dataset. Note that in some cases our method is able to pick correct segmentations that are not marked correctly in the ground truth. Best viewed in colour.

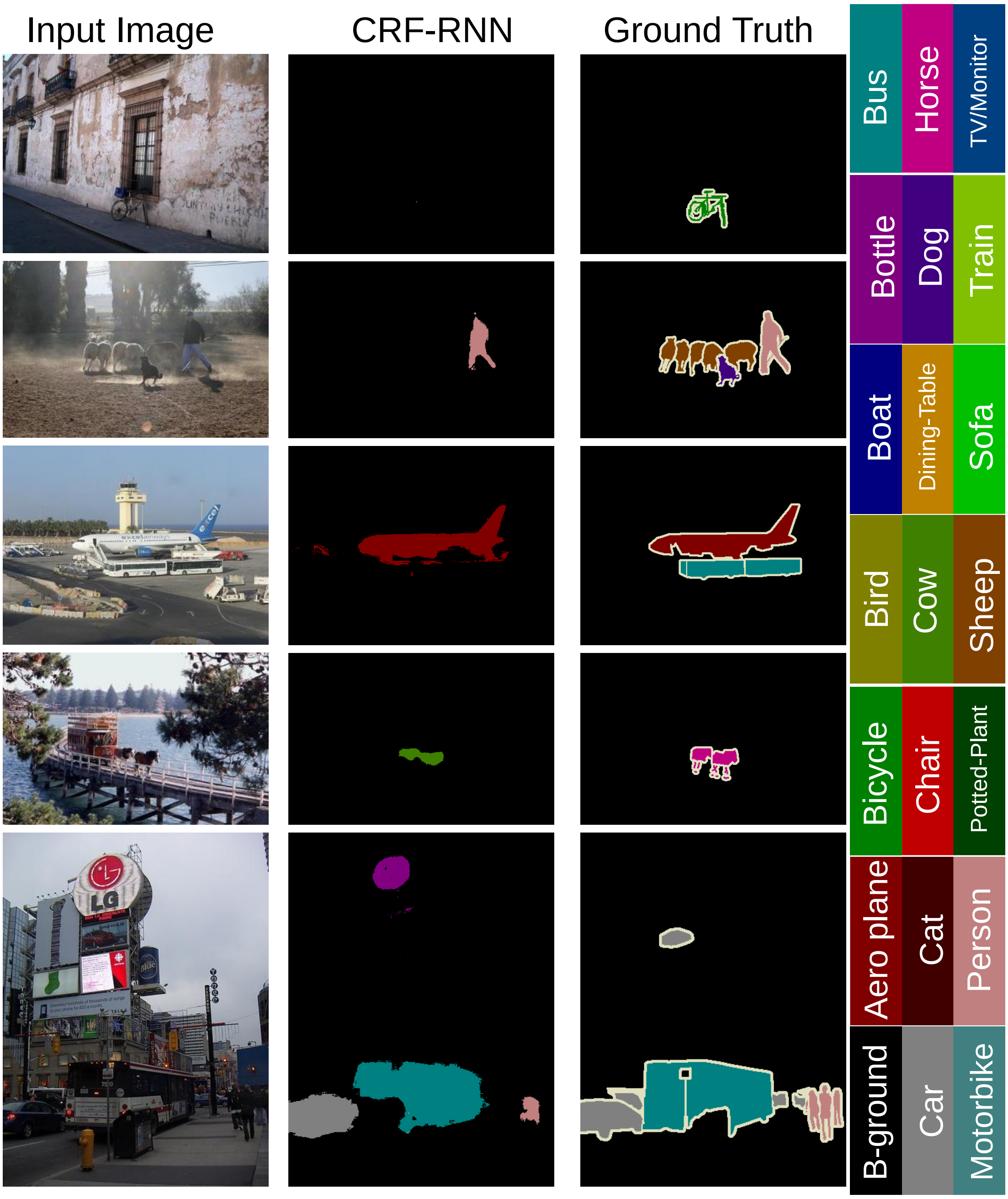

Figure 8. **Failure cases I.** Illustration of sample failure cases on the validation set of the Pascal VOC 2012 dataset. Best viewed in colour.

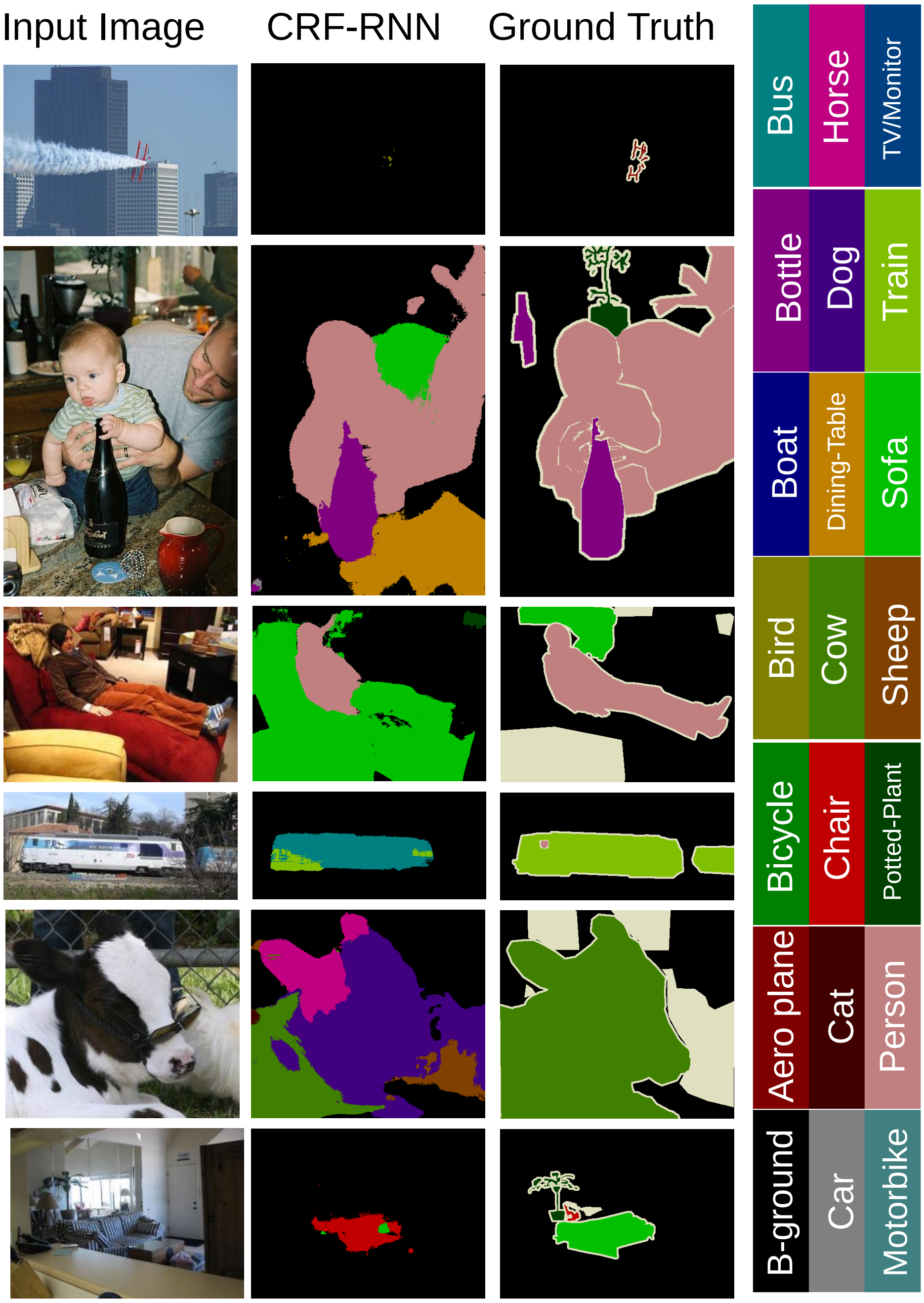

Figure 9. **Failure cases II.** Illustration of sample failure cases on the validation set of the Pascal VOC 2012 dataset. Best viewed in colour.

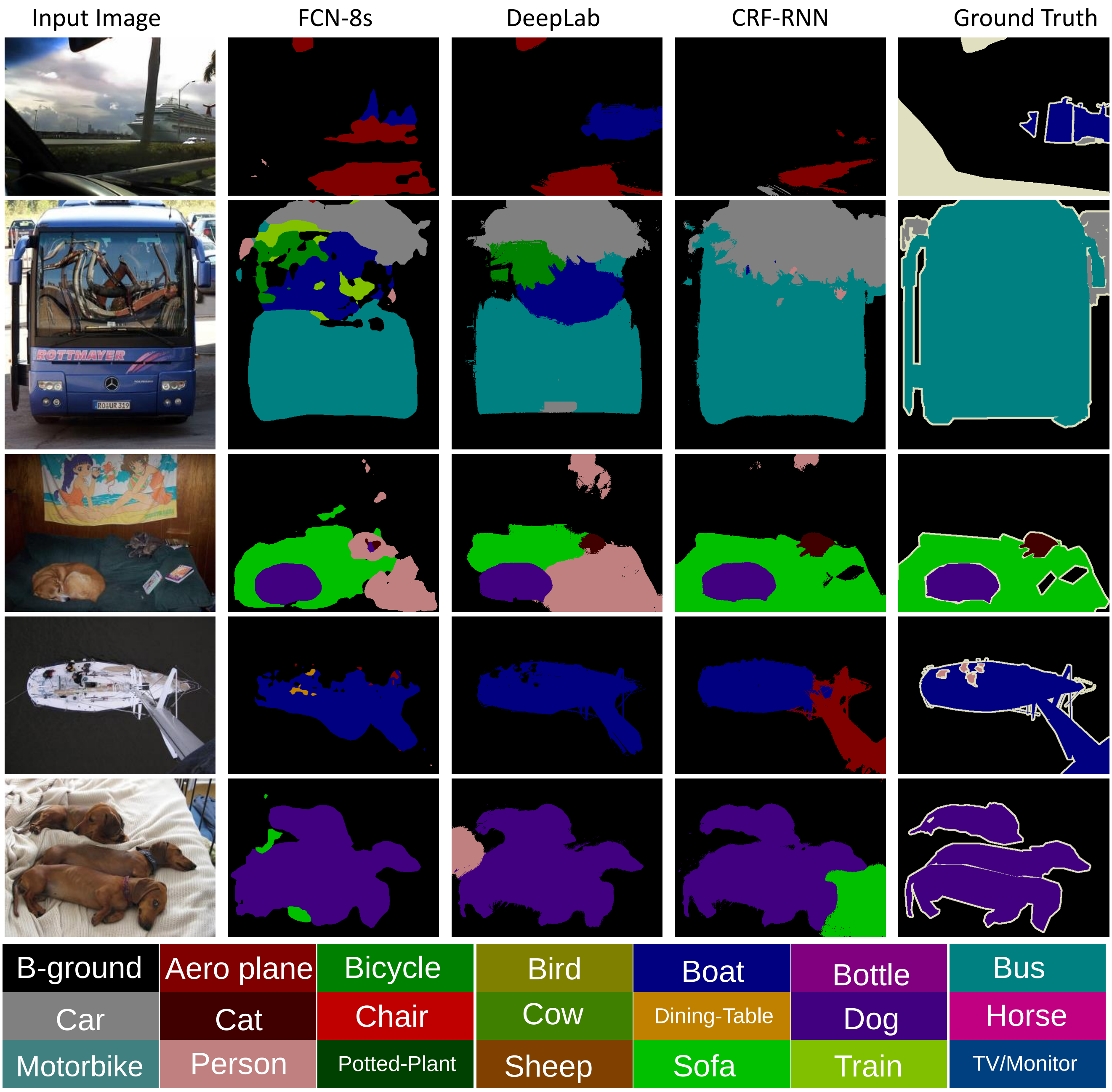

Figure 10. **Qualitative comparison with the other approaches.** Sample results with our method on the validation set of the Pascal VOC 2012 dataset, compared with previous state-of-the-art methods. Segmentation results with DeepLap approach were reproduced from the original publication. Best viewed in colour.